# Exploring Accelerated Skill Acquisition via Tandem Training for Colonoscopy

O. Richards[1], K. L. Obstein[1,2], N. Simaan[1]

[1]Department of Mechanical Engineering, Vanderbilt University, Nashville TN, USA

[2] Division of Gastroenterology, Hepatology and Nutrition,
Vanderbilt University Medical Center, Nashville TN, USA

olivia.richards@vanderbilt.edu

## INTRODUCTION

Colonoscopy training demands expert monitoring and extensive hands-on practice. Professional societies note that competency of an endoscopist should not even be assessed until at least 300 supervised procedures [1]. These training procedures introduce risk to patients and burden to experts. During a proctored colonoscopy, instruction is given through verbal direction and expert demonstration that requires tool hand-off. This tool handoff hinders skill acquisition since it inhibits the development of the necessary muscle synergies for in-hand manipulation of the colonoscope control wheels.

Dual-use training methodologies have been explored for minimally invasive surgery using systems like the da Vinci Si [2]. The benefit of having an expert-in-the-loop for training novice users include real-time haptic feedback and tool maneuver correction. The trainee is able to follow experts or lead with online expert correction. This form of in-hand, dual training has been unavailable for colonoscopy. Research dedicated to colonoscopy training has focused on developing realistic simulations [3] to improve insertion-tube control, but commonly neglect angulation control wheel steering. This creates a need for learning platforms centered on the angulation control wheels, the efficiency of which is dependent on the device remaining in the trainee's hands.

This paper investigates the feasibility of eliminating device handoff by enabling in-hand expert guidance. This is achieved through a novel concept for tandem training as shown in Fig. 1. This system allows the expert to intervene via control wheel commands using a tandem colonoscope control body (a) that is connected to an actuation unit (b) which drives the trainee's colonoscope control body (c). Using this setup, a unidirectional telemanipulation framework is used to automatically arbitrate when to pass the expert motion cues to the trainee in order to guide them during difficult maneuvers.

## MATERIALS AND METHODS

A custom preceptor control body (PCB) shown in Fig.1a was integrated with the robotic mediation device (RMD) presented in [4]. The PCB is an SLA printed standard colonoscope control body that encases two Same Sky AMT10E2-V modular incremental encoders. The encoders track the relative position of each concentric shaft of the angulation control wheels. The position of the PCB control wheels is used for telemanipulation of the RMD's adapted control body.

Our system has two operation modes: active compliance and preceptor guidance mode. The active compliance mode, the design of the RMD, and the feed-forward actuation torque commands for bending the colonoscope were presented in [4]. In this paper we introduce the preceptor guidance mode, which is triggered upon the detection of motion onset of either one of the control wheels of the PCB. The reference command for the actuation unit adjusting the trainee's control wheels is calculate using:

$$\boldsymbol{q}_i = \boldsymbol{q}_i^* + \frac{d\boldsymbol{\theta}_i}{\boldsymbol{g}_1\boldsymbol{g}_2}, \quad i = 1,2 \qquad (1)$$

where $\boldsymbol{q}_i$ is the reference motor position sent to the low-level controller of the RMD, $\boldsymbol{q}_i^*$ is the position of the motor at the onset of motion of the preceptor's corresponding control wheel, $d\boldsymbol{\theta}_i$ is the total change in angle of the preceptor's control wheel relative to its position at motion onset, and $\boldsymbol{g}_1\boldsymbol{g}_2$ are the gear ratios between the control wheel and the corresponding motor of the RMD.

A binary switch was used to toggle between active compliance and preceptor guidance modes. For this dual control scheme, we implemented a variable PID controller:

$$\boldsymbol{\tau}(t) = K_p(\sigma)\boldsymbol{e}(t) + K_d(\sigma)\dot{\boldsymbol{e}}(t) + K_i(\sigma)\int \boldsymbol{e}(t)dt$$
$$(K_p, K_d, K_i) = \begin{cases} K_{p,a}, K_{d,a}, K_{i,a}, & \text{if } \sigma = 0 \\ K_{p,t}, K_{d,t}, K_{i,t}, & \text{if } \sigma = 1 \end{cases} \qquad (2)$$

where $\sigma$ will shift the PID gains from active compliance control (e.g., $K_{x,a}$) to preceptor guidance (e.g., $K_{x,t}$) if the expert is rotating either wheel more than 0.02 degrees between control cycles (500 Hz implemented on a real-time control computer).

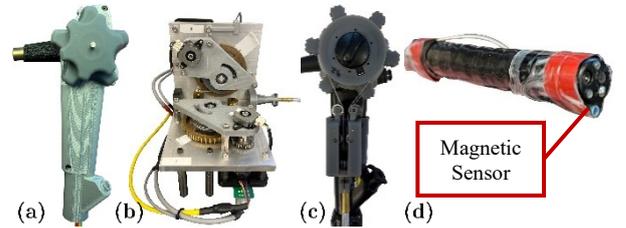

**Fig. 1.** Tandem training system: a) Custom preceptor colonoscope; b) Actuation Unit of the RMD; c) Adapted Control Body of the RMD; d) Insertion tube mounted with four 3D Guidance trakSTAR[TM] 2 Model 130 sensors

With our tandem controlled colonoscope, we performed an exploratory user study with 4 novice users and 1 expert to quantify the impact on learning rate that having an expert in-the-loop would have. All users performed 10 colonoscopies on a Kyoto Kagaku silicon colon model, which was in a normal loop configuration. The motion of the colonoscope tip and stem was tracked using an array of 4 magnetic trackers using the NDI trakSTAR[TM] 2 magnetic tracker by Ascension.

We divided the novices into two groups: one with expert intervention every other procedure, and another with no intervention. Novice users were instructed to perform a colonoscopy, but were given no instruction or coaching before or during the procedure. During tandem training, the expert would provide a verbal three-count before assuming control after three consecutive seconds of no forward advancement or poor visualization of the colonic lumen. While the expert maneuvered the colonoscope, the novice user's control wheels would follow, allowing the trainee to learn the correct wheel motions. The procedure's completion time and path of the colonoscope's tip was recorded for all users.

## RESULTS

Novice users frequently misdirected the scope into the colon's walls resulting in longer procedure times and longer path length. To visualize their improvement for navigation time, we compared the normalized completion time for each experiment performed by all users as shown in Fig.2.

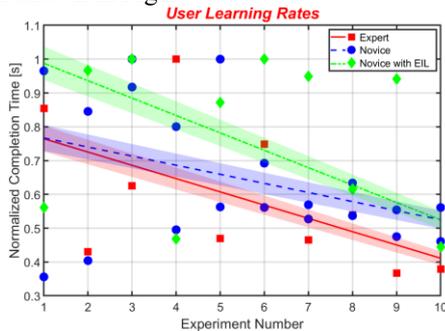

**Fig. 2** Learning rate of expert, novice, and novice with expert in-the-loop (EIL). The shaded region represents a 5% margin. Only the experiments without expert intervention were used to fit the line.

The expert's ten procedures were fitted with a line to represented an "ideal" learning rate. The novice users without expert guidance had a slower learning rate overall and the learning rate of novice users with expert guidance more closely followed the "ideal".

An example of the expert's impact on path length is shown in Fig.3. This plot shows the same section of the colon from three consecutive novice colonoscopies. The change in total path length between these trials was a) 3.02 m., b) 2.08 m., c) 2.46 m. showing that before and after the expert in-the-loop, the novice user was able to complete the procedure with 18.5% improvement in those three procedures alone.

To compare the overall effect of expert guidance for each group, we calculated the average percent improvement between the first and last unassisted colonoscopy for time and path length. From trial one to trial ten, the expert showed an improvement of 55.6% in completion time and 17.0% in total path length. The novice users without expert intervention had an average improvement of 6.2% for time and 5.0% for path length. The novice users with expert intervention had an average improvement of 53.4% in time and 14.6% in path length.

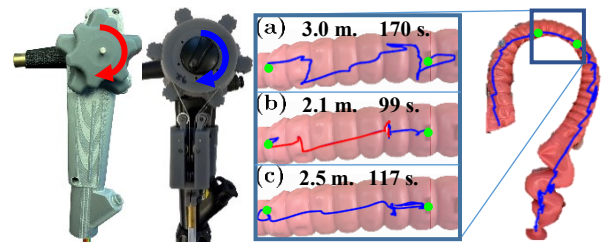

**Fig. 3** Left: Preceptor motion (red) and novice motion (blue). Center: comparison of total path length and time for a) the trial before expert intervention, b) the next trial showing the expert's correction in red, c) the following trial with corrected path. Right: example of full colonoscopy path.

## DISCUSSION

This is the first case of using telemanipulation to accelerate skill acquisition during in-hand complex finger manipulation. While this work is exploratory in nature and is limited by the small cohort size of the users, these early results suggest that accelerated skill acquisition may be possible using our system. One factor that may confound the results of every user study involving trainees is maintaining a constant pressure to motivate trainees to remain focused and attempt to do their best in each attempt. This was done via the 3-second counting rule which allowed the expert to notify the trainee that they are about to receive guidance if they are unable to produce forward motion of the colonoscope. Another limiting aspect of this work is that we only tested on a single configuration of the colon known to be the easiest. While we used the easiest colon configuration, our assumption is that the benefits shown for the easiest configuration of a normal loop would be magnified when using a more complex colon loop configuration. Future work will include adding actuation to the PCB to allow bidirectional telemanipulation. This will haptically communicate the novice's motion to the expert and enable greater insight for individualized training.

## ACKNOWLEDGEMENT

This work was supported by the VISE Seed Grant and the National Institute of Biomedical Imaging and Bioengineering (NIBIB), USA of the National Institutes of Health (NIH) under award no. 2R01EB018992.